\documentclass[lettersize,journal]{IEEEtran} 

\IEEEoverridecommandlockouts                              

\usepackage{graphicx} 
\usepackage{bm}
\usepackage{hyperref}

\usepackage{tabularx}

\usepackage{amssymb}
\usepackage{float}
\usepackage{pifont}

\usepackage{siunitx}
\usepackage{subcaption}
\usepackage{textcomp}
\usepackage[linesnumbered,ruled,vlined]{algorithm2e}
\usepackage{algpseudocode}
\usepackage{amsmath,mathtools}
\usepackage[font=footnotesize]{caption} 
\usepackage{multirow}

\title{\LARGE \bf
Adjustbot: Bio-Inspired Quadruped Robot with Adjustable Posture and Undulated Body for Challenging Terradynamic Tasks
}

\author{Saurav Kumar Dutta, Yasemin Ozkan-Aydin*
\thanks{*Corresponding author, email:yozkanay@nd.edu}
\thanks{Both authors are with the Department of Electrical Engineering, University of Notre Dame, Notre Dame, IN 46556 USA }
}

\begin{document}
\maketitle
\thispagestyle{empty}
\pagestyle{empty}

\begin{abstract}
The ability to modify morphology in response to environmental changes represents a highly advantageous feature in biological organisms, facilitating their adaptation to diverse environmental conditions. While some robots have the capability to modify their morphology by utilizing adaptive body parts, the practical implementation of morphological transformations in robotic systems is still relatively restricted. This limitation can be attributed, in part, to the intricate nature of achieving such transformations, which necessitates the integration of advanced materials, control systems, and design approaches. In nature, a range of morphology adaptation strategies is employed to achieve optimal performance and efficiency, such as those employed by crocodiles and alligators, who adjust their body posture depending on the speed and the surface they traverse on. Drawing inspiration from these biological examples, this paper introduces Adjustbot, a quadruped robot with an undulating body capable of adjusting its body posture. Its adaptive morphology allows it to traverse a wide range of terradynamically challenging surfaces and facilitates avoidance of collisions, navigation through narrow channels, obstacle traversal, and incline negotiation.
\end{abstract}
\begin{IEEEkeywords}
Variable morphology, Posture change, Lateral undulation, Terradynamics
\end{IEEEkeywords}

\section{Introduction}

A biological creature's capacity to adapt to a variety of environments is greatly facilitated by morphological changes that increase its robustness \cite{bongard2011, nygaard2021, cianchetti2015}. The creatures adopt many different kinds of morphological changes to adapt themselves to the changing environment \cite{mintchev2016}. 
Pigeons are known to use a strategy of folding their wings in order to navigate through obstacles \cite{williams2015}. Also, to fly more efficiently and conserve energy, bats, and birds constantly fold and unfold their wings \cite{riskin2012}. In addition to avoiding obstacles, several species of creatures like caterpillars \cite{brackenbury1997} and salamanders \cite{garcia1995}, coil their bodies like a wheel and roll to avoid the predators. 

\begin{figure}[!t]
\centering
\includegraphics[width=\columnwidth]{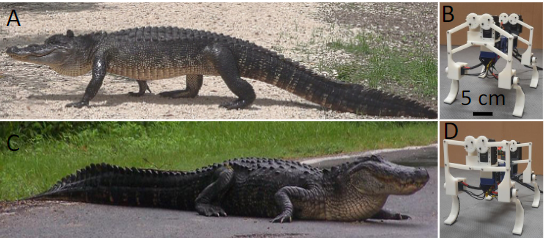}
\caption{\textbf{Semi-erect posture in} (A) crocodiles \cite{dashnau2022} and (B) our robotic model, Adjustbot. Sprawl posture in (C) crocodiles \cite{virginia} and (D) our robotic model.}
\vspace{-7mm}
\label{fig2}
\end{figure}

Morphological changes during locomotion are also observable in legged animals such as crocodiles and alligators, as these creatures exhibit the ability to modify their body posture in response to the speed and surface conditions of their locomotion. For instance, when walking at slower speeds on wet and muddy surfaces, crocodiles adopt a sprawled posture \cite{reilly1998}. Semi-erect posture is used to walk with higher speeds on dry surfaces \cite{reilly1998}. Figure \ref{fig2} shows the semi-erect and sprawl postures of a crocodile. Zoologists generally classify postures found in vertebrates into three types: sprawling, semi-erect, and erect on the basis of the sidewise relative movement of the femur from the body \cite{reilly1998}. Mammals like humans, dogs, cats, cows, and horses, to name a few have mostly erect posture \cite{reilly1998} whereas salamanders and geckos have a sprawling posture \cite{wang2018}.

\begin{figure*}[!t]
\centering
\includegraphics[width=1\textwidth]{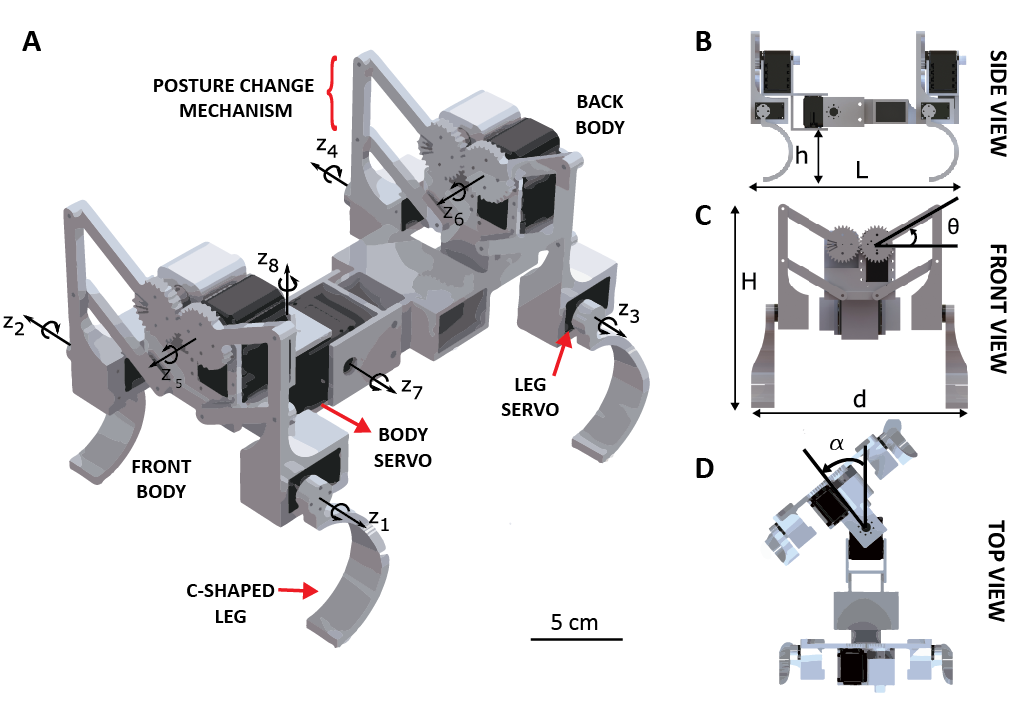}
\caption{\textbf{Mechanical parameters of the robot}. (A) An isometric view (perspective) of the 3-D CAD model showing the different parts along with the axes of rotation. $z_{1}, z_{2}, z_{3}, z_{4}$ correspond to the axes of rotation of Dynamixel XL330-M288-T, $z_{5}$ and $z_{6}$  correspond to the axes of rotation of Dynamixel AX-12A, $z_{7}$ and $z_{8}$  correspond to the axes of rotation of Dynamixel 2XL430-W250-T. (B) Side view, L: Length, h: Distance from the ground to the base. (C) Front view, showing the posture change mechanism, d: Width, $\theta$: Posture angle from the horizontal axis. (D) Top view, showing the lateral undulation, $\alpha$: Amplitude of body angle}
\label{fig3}
\end{figure*}

Similar to biological systems, many researchers have developed robots with adaptive and re-configurable body parts to enhance their functionalities \cite{derrouaoui2021, kim2007, kim20212,chou2013,li2020,ozkan2021self,song2022,zheng2022,Zarrouk2023,DeforMoBot2023}. Moreover, adaptive behavior leads to the simplification of control \cite{zambrano2014} and robot design \cite{kim2013}. The mobile robot of Li et al. \cite{li2020} has a transformable body that allows it to move both in open areas and narrow channels. Jiang et al. \cite{jiang2019} make use of a balance-rocker mechanism which helps the robot to change its posture and adapt to different terrain conditions. Similarly, the six-wheeled mobile robot of Song et al. \cite{song2022} makes use of a Sarrus-variant mechanism to reconfigure the robot body for terrain adaptability and obstacle crossing. Zheng et al. \cite{zheng2022} use an adaptive wheel for multi-terrain locomotion. In the robot of Chou and Yang \cite{chou2013}, the robot body folds and unfolds itself to switch between wheel mode and claw mode. The amphibious robot of Liang et al. \cite{liang2012} uses an adaptive propulsion mechanism wherein the mechanism switches between leg and flipper to move on the surface and underwater, respectively. Baines et al. \cite{baines2020} use a morphing limb in their bio-inspired amphibious robot. The morphing limb transitions between a flipper for swimming and a leg for walking. Unmanned aerial systems are also increasingly becoming transformable as they enhance the maneuverability and agility of such systems \cite{derrouaoui2021}. Some robots also change their posture to navigate through confined spaces \cite{buchanan2019,DeforMoBot2023}. Certain quadruped robots have a bio-inspired laterally undulating adaptive spine or backbone, which helps them not just in their locomotion \cite{wang2018, chong2021, qiu2023} but also for energy-efficient climbing \cite{haomachai2021}.  

As variable morphology enhances the functionalities of both biological and engineering systems, the present paper comes up with an adjustable posture and laterally undulating bio-inspired quadruped robot, Adjustbot for terradynamically demanding situations. The majority of currently available quadruped robots \cite{sprowitz2013, kalakrishnan2011, seok2014, saranli2001, gor2018} lack bio-inspired features like changeable posture \cite{reilly1998} or lateral undulation \cite{wang2018} and have stiff bodies, which prevent them from becoming more robust in accordance with biological notions. Previously, there have been posture-changing quadruped robots, like in the work of Bongard \cite{bongard2011}, Ansari et al. \cite{ansari2015}, and, Juárez-Campos et al. \cite{ juarez2018}. However, the robot developed by Bongard \cite{bongard2011} uses a series of gears in a gear train to actuate the posture change, and also it uses legs that act as stands. The quadruped robot of Ansari et al. \cite{ansari2015} uses soft legs for posture change. Juárez-Campos et al. \cite{ juarez2018} make use of Peaucellier–Lipkin mechanism in their robot to effect posture change. The performance of these robots is unknown as they were not tested extensively in unstructured environments.

Adjustbot, the robot under study, is a three-dimensional (3D) printed, quadruped robot equipped with seven servo motors, each serving a distinct purpose (See Fig.\ref{fig2} B, D, and Fig.\ref{fig3}). The rotational motion of its C-shaped legs is facilitated by four servo motors. Additionally, two servo motors, in conjunction with a gear drive mechanism, enable posture-changing capabilities, while one servo motor facilitates lateral undulation of the robot's body.

The research presented in this paper represents a significant contribution to the field, as it highlights and substantiates, through experimental validation, the crucial role of morphology change in bio-inspired quadruped robots when navigating challenging, unstructured environments. By allowing for dynamic posture adjustments, these robots demonstrate a higher capacity to adapt to unfamiliar surroundings, making them particularly suitable for applications such as search and rescue operations during natural disasters.

The rest of the letter is organized as follows: Section \ref{secmm} provides a summary of the components and procedures utilized to design, manufacture, and control the robot. Section \ref{secstabil} discusses stability and gait selection through bio-inspiration. The experimental results are discussed in Section \ref{secexp}. Finally, Section \ref{secconc} concludes the letter.

\section{Materials and Methods}
\label{secmm}
\subsection{Robot Modeling and Design}

The 3D computer-aided design (CAD) model of the robot is shown in Fig. \ref{fig3}. The robot has eight axes of rotation, which are denoted by $z_{i, i=1,2,3,...,8}$. $z_1$, $z_2$, $z_3$, and $z_4$ correspond to the axes of rotation of the C-shaped legs. $z_{5}$ and $z_{6}$ correspond to the axes of rotation of the two posture change mechanisms that are located at the front and back segments of the robot. $z_{7}$ and $z_{8}$ correspond to the pitch and yaw motions of the robot body, respectively. The yaw motion corresponds to the lateral undulation of the robot body. It may be noted that $z_{5}$ and $z_{6}$ correspond to a single degree of freedom as the two posture change mechanisms do not have motion relative to each other. 

However, it is important to highlight that, in the context of this study, the robot's body pitch motion is not utilized. The motor corresponding to the $z_7$ axis is initialized at a zero-degree angle, as indicated in Fig. \ref{fig3}, and remains fixed at zero degrees throughout the robot's operation. Consequently, the robot has six degrees of freedom, with four associated with the C-shaped legs, one with the two posture change mechanisms, and one with the lateral undulation of the robot.

The posture change mechanism consists of two parallelogram mechanisms \cite{li2010} which are synchronously operated by a single motor with the help of a gear drive. The opposite links in the parallelogram mechanism are of the same length. In the parallelogram, the length of the link with gear is the longest (60 \si{\milli\meter}). The length of the portion of the link between the two longest links is 44.5 \si{\milli\meter} and the entire length of this link is 104.5 \si{\milli\meter}. For the gear drive, both the gears have a pitch circle diameter of 15 \si{\milli\meter} and a module (the ratio of the pitch diameter of the gear to its number of teeth) of 1 \cite{uicker2003}. The range of the posture angle, $\theta$ is between +40\si{\degree} and -60\si{\degree}. 

\begin{figure}[!t]
\centering
\includegraphics[width=0.5\textwidth]{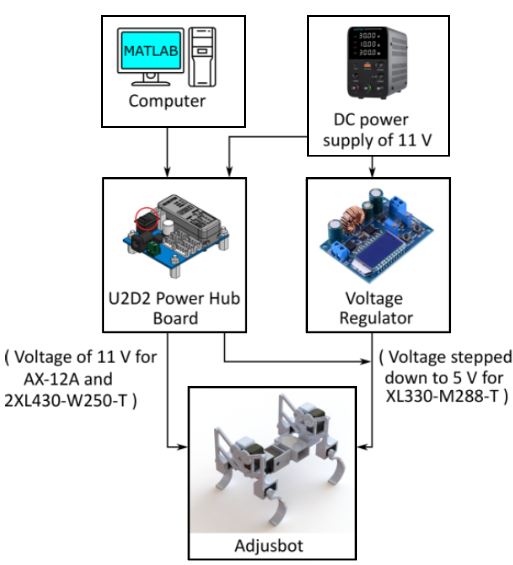}
\caption{\textbf{System architecture of the robot}: The robot is controlled using U2D2, assembled on a U2D2 power hub board, by a program written in MATLAB. A voltage regulator is used to step down the voltage from 11 to 5 for the motors driving the C-shaped legs.}
\label{fig5}
\end{figure}

\begin{figure*}[!t]
\centering
\includegraphics[scale=0.7]{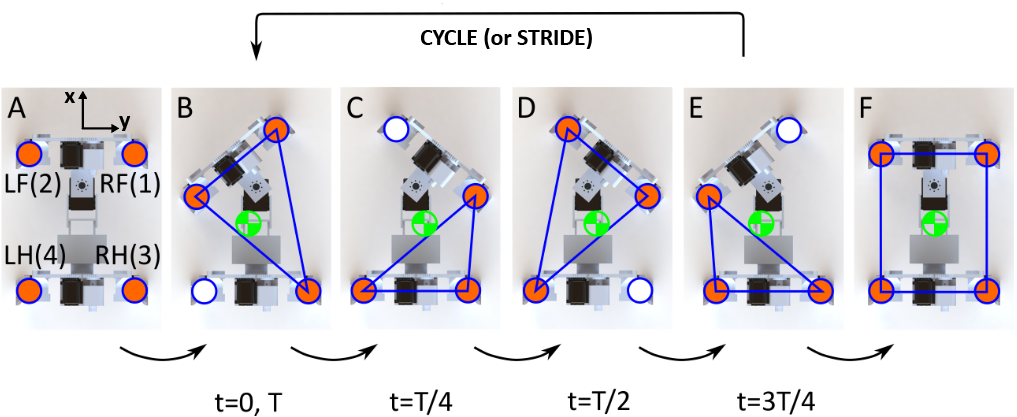}
\caption{\textbf{Robot gait pattern}: (A) The robot body rotates by an angle of +40\si{\degree}. (B) (Beginning of the cycle) The left hind leg leaves the ground at t=0 and completes one revolution in t=T/4. (C) The left fore leg leaves the ground at t=T/4 to complete one revolution and at the same time, the body also undulates from its previous position to a current position of -80\si{\degree} in t=T/2. (D) The right hind leg leaves the ground at t=T/2 and completes one revolution in t=3T/4. (E) (End of the cycle) The right foreleg leaves the ground at t=3T/4 to complete one revolution and at the same time, the body also undulates from its previous position to a current position of 80\si{\degree} in t=T. If the robot is run for more than 1 cycle, the steps keep repeating from (B) to (E). (F) The robot body rotates by an angle of -40\si{\degree} to come back to its start position}
\label{fig7}
\end{figure*}

The parts of the robot have been modeled in Solidworks 2022. The two most important design parameters of the robot are: d and H, as indicated in Fig. \ref{fig3} C. d and H depend on the dimensions of the links in the parallelogram mechanism and the posture angle, $\theta$. Hence, the dimensions of the links in the parallelogram mechanism have been chosen in a way that d and H have a range of 30 \si{\milli\meter} and 60 \si{\milli\meter}, respectively, as indicated in Table \ref{tab:1}. The selected dimensions of the links in the posture change mechanism also ensure that $\theta$ has a range of 100 \si{\degree}. The dimensions of the links, including the range of posture change mechanisms for the smooth functioning of the robot without any interference between the links and the parts, have been selected by using GeoGebra and ``animation, basic motion, and motion analysis" options in Solidworks 2022. Similarly, while designing the body of the robot, it is ensured that the amplitude of the body angle, $\alpha$ has a range of at least 60\si{\degree} for each of the postures of the robot. It may be noted that the maximum $\alpha$ for each of the 40, 0, and -60 degree postures of the robot without any interference between the limbs and the body are 70, 60, and 75 degrees, respectively. The robot has C-shaped legs as these legs provide a fair trade-off between the enhanced mobility of legged robots and the sturdy stability of wheels \cite{altendorfer2001rhex,vina2021}. The inner and outer diameter of the C-shaped legs are 30\si{\milli\meter} and 35\si{\milli\meter}, respectively. These legs make sure that h has a minimum value of 25 \si{\milli\meter} at $\theta$ = +40\si{\degree} posture. The front body of the robot, from $z_{8}$ axis to the front face along the longitudinal axis is 70\si{\milli\meter} long. Thus, the front body of the robot rotates by an angle $\alpha$ with a radius of 7070\si{\milli\meter}. The dimension has been chosen such that the robot has at least an $\alpha$ of $\pm{60\si{\degree}}$ for all the possible postures of the robot.   

While a power source is employed for powering the robot, we also integrated a battery box into the central section of the robot, as illustrated in Figure \ref{fig3}, with the intention of facilitating future autonomous locomotion capabilities.

\begin{table}[!t]
\centering
\caption{Design parameters of the robot (Refer Fig. \ref{fig3})}
\label{t4}

\begin{tabular}{ c c c c c }
\hline
\hline
Posture &  Width  & Height &  Length & Height from ground \\
$\theta$ & d & H & L & h \\
\hline 
40\si{\degree} &  180 & 180 & 250 & 25    \\

   &  &  &    \\

0\si{\degree} &  202 & 190 & 250 & 65 \\

   &    &    &     \\

-60\si{\degree} & 150  & 240 & 250 & 110  \\
\hline
\end{tabular}
\label{tab:1}
\end{table}

\subsection{Prototype Development}
After the design process and finalizing the model of the robot, its parts are fabricated using 3D printing in Stratasys F-170 printer with an Acrylonitrile Butadiene Styrene (ABS) plastic material.  The final prototype of the robot is shown in Fig. \ref{fig2}. In order to increase the friction between the leg and the ground, the backside of all the legs is covered by a thin patch (3 \si{\milli\meter}) of rubber. 

\subsection{Actuators and Electronics Integration} 

 The robot uses a total of seven actuators, four Dynamixel XL330-M288-T (0.52 [N.m], 5V) motors for the legs, two Dynamixel AX-12A (1.5 [N.m], 11.1V) for the two posture change mechanisms and one Dynamixel 2XL430-W250-T (1.4 [N.m], 11.1V)for the lateral undulation. Each of the motors is independently controlled using Robotis U2D2. U2D2 is assembled in the U2D2 Power Hub board, which is powered by a DC supply of 11 \si{\volt}. A program is written in Matlab R2022b and is fed to U2D2 in real time to control each of the axes of the robot independently. The system architecture of the robot is shown in Fig. \ref{fig5}. The entire control system is open-loop. In the system architecture, a voltage regulator is also used as the recommended voltage of Dynamixel XL330-M288-T is 5 \si{\volt}.  

The control program for the robot is written in MATLAB 2022 using the Robotis Dynamixel SDK Matlab library \cite{DynamixelSDK}. In this study, we have configured the operating mode of the leg and body servos to utilize the extended position control mode. This mode enables the legs to rotate beyond 2$\pi$ in either direction and facilitates the synchronous movement of the robot body with the legs. Regarding the two posture change mechanisms, we have employed the position control mode for both motors. This allows precise control over the positioning of the mechanisms throughout the robot's operation.

\section{Stability and Gait Selection}
\label{secstabil}

To maintain static stability throughout the robot's walking motion, it is essential to ensure that the vertical projection of the robot's center of gravity consistently remains within the support pattern (a polygon formed by connecting the leg contact points of the robot\cite{mcghee1968}). 

In the context of gait patterns, a lateral sequence gait is characterized by the hind foot of a quadruped making initial ground contact, followed by the forefoot on the same side of the body, according to Hildebrand \cite{hildebrand1965}. Meanwhile, a creeping gait implies that a minimum of three feet maintain contact with the ground consistently \cite{mcghee1968}. 

\begin{figure}[!t]
\centering
\includegraphics[width=0.5\textwidth]{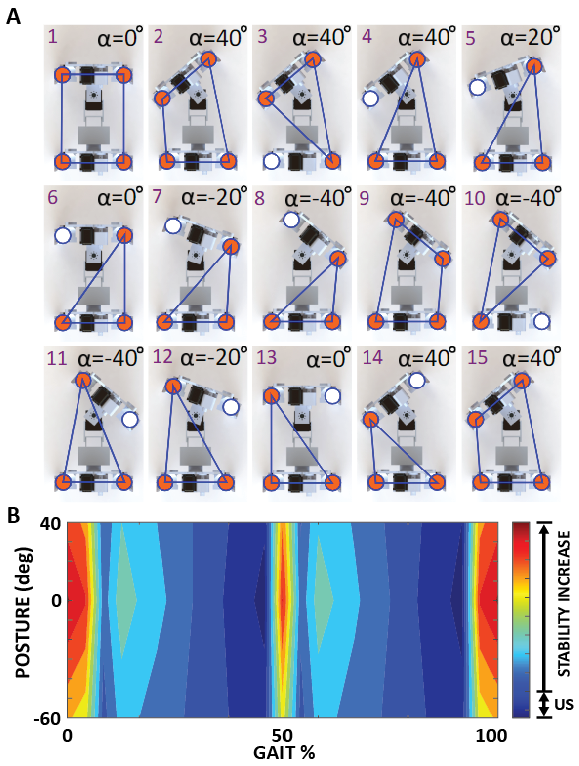}
\caption{\textbf{Stability Map}: (A) Important steps in the motion of the robot from the beginning to the end of one cycle. If the robot is run for multiple cycles, it starts from Step 1 and then keeps repeating the gait cycle between Steps 3 and 14. Orange dots represent one of the binary states when the robot leg is on the ground, while the white dot represents the leg in the air. The blue polygon represents the support polygon or triangle. (B) Stability map of the robot as a function of the gait cycle for posture angles from -60\si{\degree} to 40\si{\degree}. The colorbar indicates the stability margin values. The robot is unstable (US) when the stability margin is negative. }
\label{fig20}
\end{figure}
In this paper, we adopted the lateral sequence creeping gait due to its capacity to facilitate low-speed locomotion while ensuring stability for the quadruped throughout the majority of the locomotion cycle (Fig. \ref{fig7}). This choice stems from the fact that a creeping gait, as described in \cite{tomovic1961}, guarantees a positive static stability margin across the majority of the gait cycle \cite{mcghee1968}. Furthermore, we selected the lateral sequence gait \cite{hildebrand1965} in light of its advantages: (1) the area enclosed by the support triangle is greater than that enclosed by the support triangle for a diagonal sequence gait \cite{gray1944} and (2) a lateral sequence gait helps in the lateral undulation of a quadruped \cite{hildebrand1980}. 

Solidworks is used to validate the stability of the robot, incorporating a lateral sequence creeping gait in conjunction with a laterally undulating back. The stability margins of the robot for different body angles, $\alpha$, and posture angles, $\theta$, during a cycle are obtained using Solidworks.  These values are then utilized to generate the contour plot depicted in  Fig. \ref{fig20}. For all the postures, the stability margins are the greatest for steps 1, 2, 9, and 15, when the four legs touch the ground. The stability margins are positive for all the steps except for steps 7, 8, and 14. During these steps, the foreleg remains in the air and the robot has a tendency to fall in the forward direction. However, before the robot body could fall and touch the ground, the foreleg completes its rotation and touches the ground first. Due to this, the robot regains its stability. From Fig. \ref{fig20}, it can be seen that in the immediate step after steps 8 and 14, the four legs maintain contact with the ground and the stability margin becomes greatly positive from negative.

\begin{figure}[!t]
\centering
\includegraphics[width=\columnwidth]{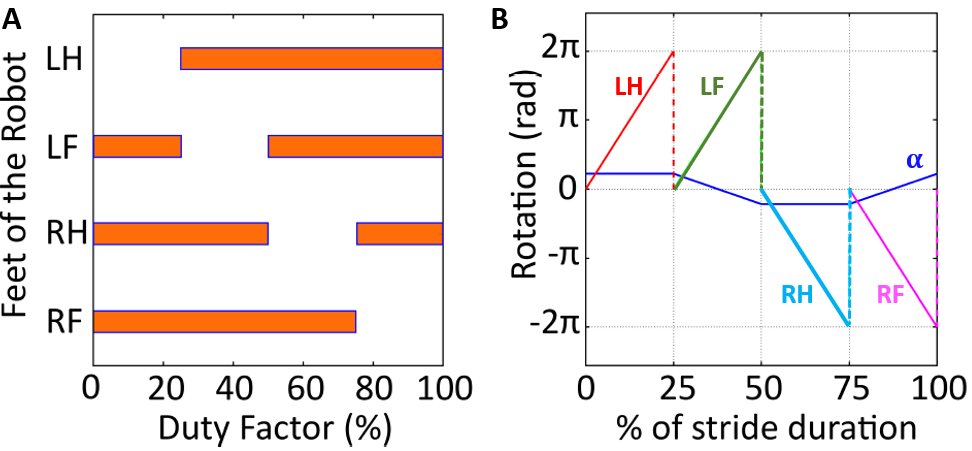}
\caption{\textbf{Gait parameters}: (A) Gait diagram. LH: Left hind foot. LF: Left forefoot. RH: Right hind foot. RF: Right forefoot.  The orange bar denotes the percent of stride interval that each foot is on the ground (B) The sequence of motion of the legs and the back. $\alpha$: Body angle.}
\label{fig6}
\end{figure}

The gait diagram \cite{hildebrand1989} and the sequence of motion of the legs and the body angle for the lateral sequence gait are shown in Fig. \ref{fig6}. From Figs. \ref{fig7} and \ref{fig6}, it can be seen that each of the legs remains in the air for the same amount of time (25\%) during a cycle while the other three legs are in contact with the ground. Hence, according to Hildebrand's gait formula, the robot uses a symmetrical gait and has a gait formula of (75,25) \cite{hildebrand1965}. Here, the first variable corresponds to the percentage of time that each hind foot is on the ground during a cycle or stride (duty cycle) and the second variable corresponds to the percentage of time when a forefoot falls on the ground behind a hind foot on the same side of the body during a stride (relative phase) \cite{hildebrand1965,mcghee1968a}. McGhee \cite{mcghee1968a} generalizes Hildebrand's gait formula to include the duty factor of each of the legs of a quadruped and the relative phases of any three legs, with respect to a chosen fourth leg. According to McGhee's gait formula \cite{mcghee1968a}, $\bm{g}$ our gait is given as \vspace{-3mm}
\begin{align}
g=(0.75, 0.75, 0.75, 0.75, 0.5, 0.75, 0.25)^{T} \vspace{-3mm}  
\label{eq1}
\end{align}
In the above formula, the first four variables correspond to the duty factors of the first, second, third, and fourth legs. The remaining three variables correspond to the relative phases of the second, third, and fourth legs with respect to the first leg. This gait formula can be calculated from the gait matrix, $\bm{G}$, and the duration vector, $\bm{t}$ \cite{mcghee1968a} for the robot. For writing $\bm{G}$ and $\bm{t}$, the legs have been numbered as shown in Fig. \ref{fig7}. The matrix $\bm{G}$ comprises four columns, representing the four legs of the robot, while the total number of rows is equal to the length of one cycle of the gait sequence\cite{mcghee1968a}.  The values within $\bm{G}$ are binary, with 0 indicating a leg in contact with the ground and 1 indicating a leg lifted into the air \cite{mcghee1968a}. 

\vspace{-2mm}
\begin{equation}
\label{eq2}
\bm{G} =
\left [
\begin{array}{cccc}
0 & 0 & 0 & 0\\
0 & 0 & 0 & 1\\
0 & 0 & 0 & 0\\
0 & 1 & 0 & 0\\
0 & 0 & 0 & 0\\
0 & 0 & 1 & 0\\
0 & 0 & 0 & 0\\
1 & 0 & 0 & 0
\end{array}
\right], \;
\bm{t} =
\left \{
\begin{array}{c}
0 \\ T/4 \\ 0 \\ T/4 \\ 0 \\T/4 \\ 0 \\T/4
\end{array}
\right\}
\end{equation}

in Eq. \ref{eq2}, T is the stride duration.

The gait formula in Eq. \ref{eq1} gives information only about the duty factor and relative phases of the legs. A quadruped robot's supporting feet position and vertical center of gravity projection, along with its duty factor and relative phases, are all completely specified by a kinematic gait formula \cite{mcghee1968}. The kinematic gait formula of the robot for $\theta$ = 40\si{\degree}, 0\si{\degree} and, -60\si{\degree} with $\alpha$ = 0\si{\degree} for each of them are given as-

\begin{align}
\begin{split}
\label{eq3}
k_{40\si{\degree}}=&(0.75, 0.75, 0.75, 0.75, 0.681, 0.681, -0.509, -0.509, \\
& 0.479, -0.472, 0.479, -0.472, 0.5, 0.75, 0.25)^{T} \\    
k_{0\si{\degree}}=&(0.75, 0.75, 0.75, 0.75, 0.564, 0.564, -0.421, -0.421, \\
& 0.467, -0.467, 0.467, -0.467, 0.5, 0.75, 0.25)^{T}    \\
k_{-60\si{\degree}}=&(0.75, 0.75, 0.75, 0.75, 0.509, 0.509, -0.381, -0.381, \\
& 0.289, -0.284, 0.289, -0.284, 0.5, 0.75, 0.25)^{T}  
\end{split}
\end{align}

In the above formula, the first four and the last three variables denote the duty factors and the relative phases of the legs, respectively. The fifth, sixth, seventh, and eighth variables correspond to the x-coordinate, and the ninth, tenth, eleventh, and twelfth variables correspond to the y-coordinate of the dimensionless initial foot position of the legs of the robot. The origin lies at the center of gravity of the robot with the x-axis pointing towards the direction of motion of the robot, and the y-axis is pointing ninety degrees to the left of the x-axis. The scale of the x and y coordinate axes is chosen such that stride length is equal to 1 \cite{mcghee1968}.

As the duty factor of each of the legs is the same (0.75), which is also evident from Fig. 6 A, the gait of the robot would be represented by Hildebrand's gait formula (75,25) throughout the text. The kinematic gait formula gives information about the foot positions and the stride length of the robot, which can be determined only after the robot completes one cycle of motion. The motion Analysis option in Solidworks was used to determine the initial foot positions and center of gravity for determining the kinematic gait formula (\ref{eq3}) for the three postures of the robot. It may be noted that throughout the text, the robot makes use of only one gait. But there are different ways of representing the same gait. The robot here has a symmetrical gait and each of the legs has the same duty factor, hence Hildebrand's representation of the gait formula is sufficient with just two variables. The kinematic gait formula is much more informative but it can be calculated only after knowing the robot's initial foot positions and stride length, which vary for different postures. But across all the postures, the robot's gait formula with Hildebrand's representation remains fixed at (75,25).         

\begin{figure}[!t]
\centering
\includegraphics[width=\columnwidth]{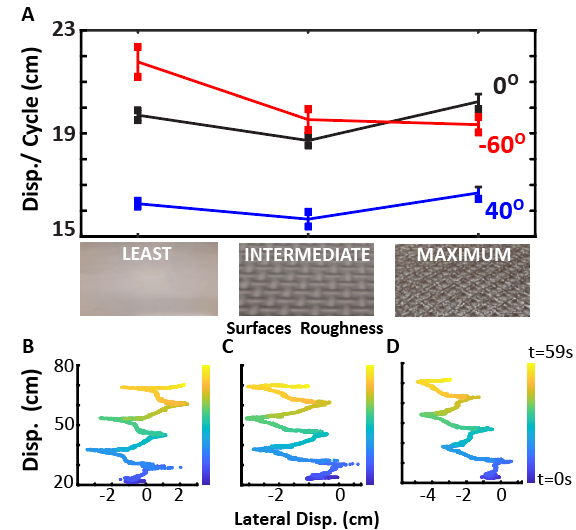}
\caption{\textbf{Walking parameters of the robot}: (A) Mean$\pm$SD of the
CoM displacement per cycle for different postures (-60$^\circ$,0$^\circ$,40$^\circ$) on surfaces of three different roughnesses (five trials and three cycles per trial). Example CoM displacement of the robot with 40$^\circ$ posture for three cycles when it walks on the surface with (B) least, (C) intermediate, and, (D) maximum roughness. The color bar indicates the time.}
\label{fig14}
\end{figure}

\section{Experimental Results}
\label{secexp}

The robot's performance in terms of overcoming hurdles without falling (robot body striking the ground) in different unstructured environments is carried out with a gait formula of (75,25) and an amplitude of body angle, $\alpha$ = 40\si{\degree} except for the ramp experiments.   

\subsection{Flat surface}

The robot's locomotion was assessed across diverse postures during three walking cycles on distinct flat surfaces, each featuring varying levels of roughness (Fig. \ref{fig14}). These surfaces encompass a whiteboard, characterized by minimal roughness, a foam mat with intermediate roughness, and a spiked plastic plate presenting the highest degree of roughness. Surface roughness levels were ascertained through visual inspection.

\begin{figure}[!t]
\centering
\includegraphics[width=0.8\columnwidth]{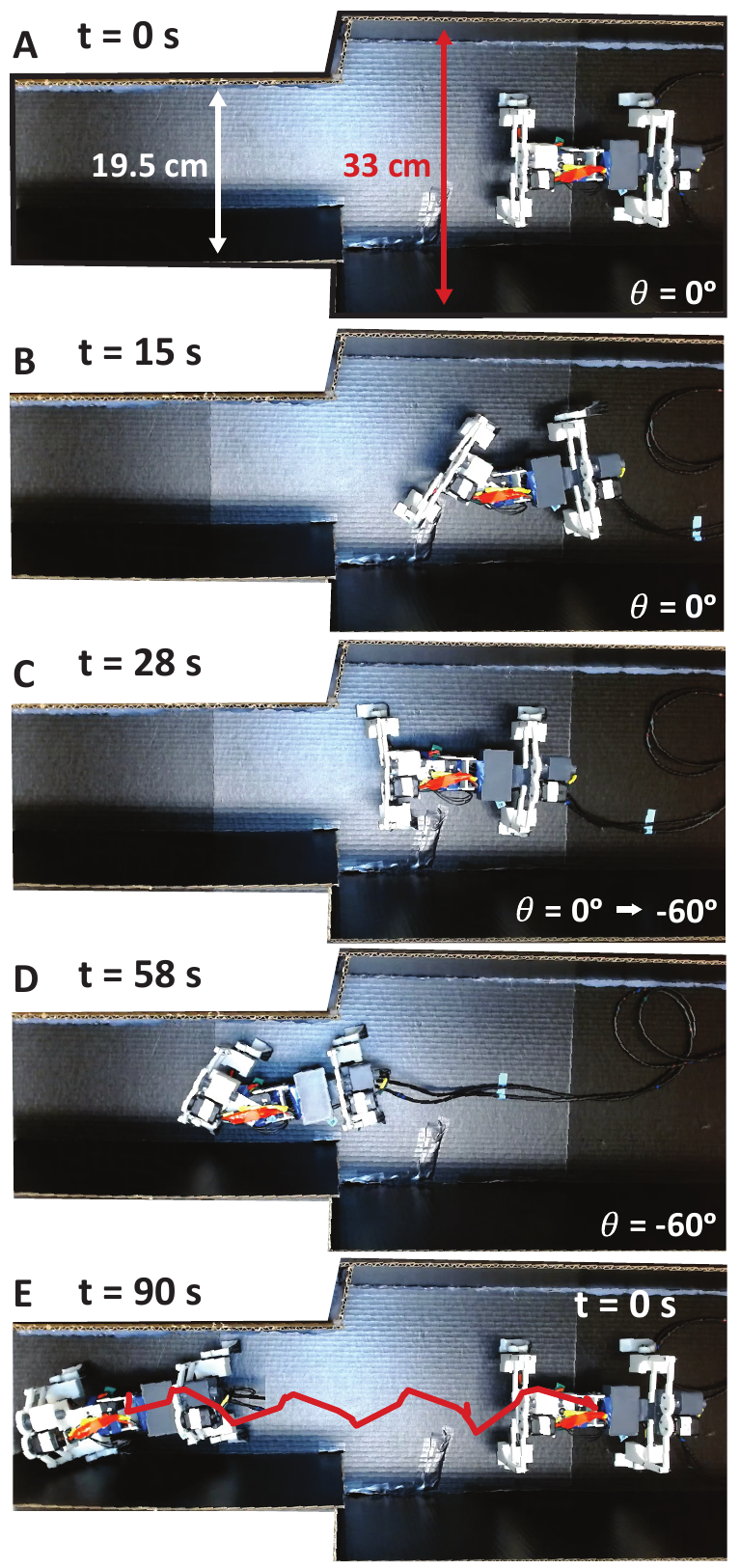}
\caption{\textbf{Moving from a broader (width=330\si{\milli\meter}) to a narrower channel (Width=195\si{\milli\meter}) through posture change}: (A) At t=0 \si{\second} with 0\si{\degree} posture. (B) At t=15 \si{\second} with 0\si{\degree} posture. (C) Posture change at t=28 \si{\second}. (D) At t=58\si{\second} with -60\si{\degree} posture. (E) Trajectory tracking of the center of the robot from t=0\si{\second} to t=90\si{\second}.}
\label{fig9}
\end{figure}
The robot's center of mass (CoM) motion was captured using the OptiTrack Motion Capture System for three specific postures: 40\si{\degree}, 0\si{\degree}, and -60\si{\degree} on each of the three surfaces. These three postures are chosen because as mentioned in Section \ref{secmm}, the posture angle, $\theta$ has minimum and maximum values of -60\si{\degree} and 40\si{\degree}, respectively. The robot can attain any posture between these two values. For analysis, an intermediate posture of 0\si{\degree} is also chosen. The motion of the robot was recorded for five trials starting from the same initial position. The mean and standard deviation (SD) values of the captured motion are shown in Fig. \ref{fig14} A. For -60\si{\degree} posture, the displacement per cycle of the robot, as shown in Fig. \ref{fig14} A is highest when it moves on the surface with the least roughness, followed by the surfaces with intermediate and maximum roughness. This is because, during the robot's locomotion, the hind legs mostly slide forward (SI movie- 0:33 to 0:44). This sliding is smooth and fast if the surface is smooth. As the surface gets rough, the hind legs tend to get stuck in the mechanical asperities as they move forward and hence the sliding is not smooth and fast. Similar to -60\si{\degree} posture, the displacement per cycle follows a pattern for 40\si{\degree} and 0\si{\degree} postures, when the robot moves on surfaces with least and intermediate roughness, as shown in Fig. \ref{fig14} A. However, contrary to -60\si{\degree} posture, the robot's displacement per cycle for the 40\si{\degree} and 0\si{\degree} postures becomes the greatest when it moves on the surface with maximum roughness. This is because during the locomotion, when the hind legs get stuck, another phenomenon happens. The robot body tends to fall forward and the suspended foreleg touches the ground before it is able to complete the full 2$\pi$ revolution (SI movie- 0:08 to 0:32). When this happens, the robot covers a greater distance in a single cycle.

Ideally, the displacement per cycle of the robot, irrespective of the surface it walks on, should have been the highest for 0\si{\degree} posture followed by 40\si{\degree} and -60\si{\degree} postures. This is because, at 0\si{\degree} posture, d (Table \ref{tab:1}) is highest, hence the arc length subtended by the front body of the robot during lateral undulation is greatest and it covers a greater distance. At -60\si{\degree} posture, d (Table \ref{tab:1}) is least and hence the arc length subtended by the body servo is less.    

Thus, three cases may arise when the hind legs of the robot tend to get stuck, which is why the displacements don't match up with the ideal displacements. In the first case, the hind leg will get past that asperity after getting stuck initially but this will lead to a decrease in displacement per cycle. In the second case, the robot will tend to fall forward and the foreleg will touch the ground before it completes a full 2$\pi$ revolution. In the third case, the robot will deviate from its straight line path and its trajectory will bend, which will contribute to the robot's lateral displacement, leading to a reduction in displacement per cycle of the robot, as shown in Figs. \ref{fig14} B,C, and D. The robot's locomotion may fall under any of these three cases and hence the displacement per cycle will vary.

\begin{figure}[!t]
\centering
\includegraphics[width=\columnwidth]{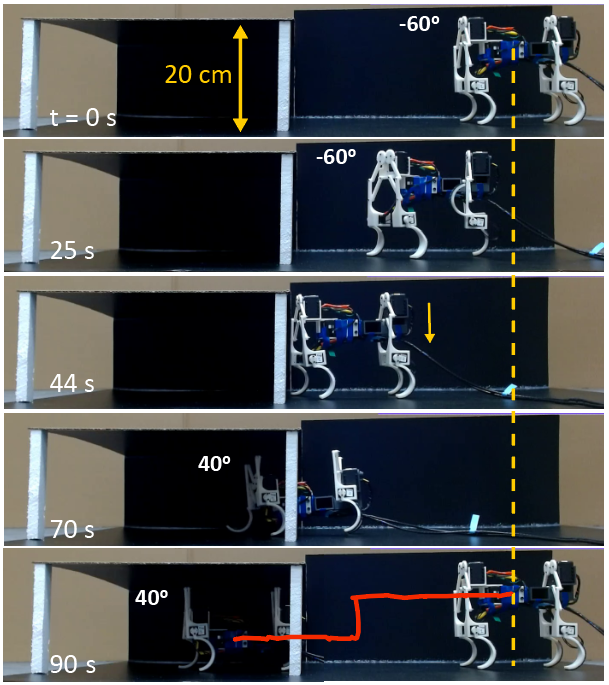}
\caption{\textbf{Moving from an open area to a tunnel (Height=20\si{\centi\meter}) through posture change}. At t=0\si{\second} and t=25\si{\second} walks with -60\si{\degree} posture. At t=44\si{\second} it changes the posture from -60\si{\degree} to 40\si{\degree}. At t=70\si{\second} and t=90\si{\second} it walks with 40\si{\degree} posture. The red trajectory shows the movement of the center of the robot from t=0\si{\second} to t=90\si{\second}}.
\label{fig10}
\end{figure}
       
\subsection{Narrow Tunnel Traversal}

\begin{figure}[!t]
\centering
\includegraphics[width=\columnwidth]{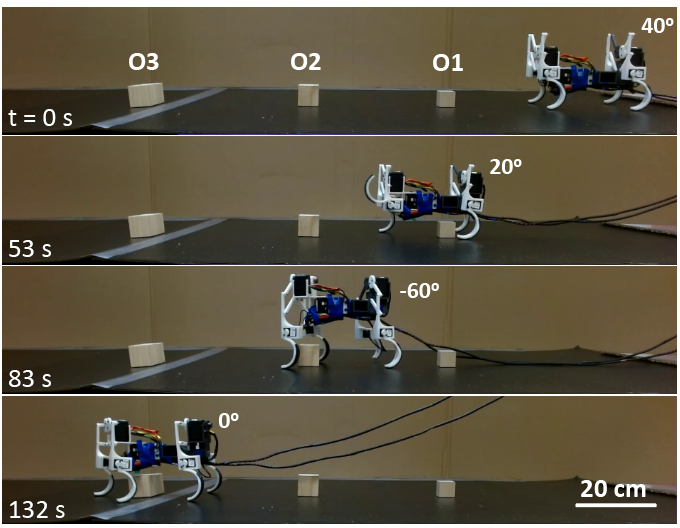}
\caption{\textbf{Moving over the different size of obstacles through posture change}:  At t= 0 \si{\second} with 40 \si{\degree} posture, at t= 53 \si{\second} it passes the first obstacle (O1= 38x38x38 mm$^3$) with 20 \si{\degree} posture, at t= 83 \si{\second} it passes the second obstacle (O2 = 50x50x50 mm$^3$) with -40\si{\degree} posture, and at t= 132 \si{\second} it passes the third obstacle (O3 = 100x50x50 mm$^3$) with 0\si{\degree} posture. }
\label{fig11}
\end{figure}
Several trials were conducted to assess the robot's ability to navigate through different widths of channels (Fig. \ref{fig9})  without and with adjusting its posture. As shown in Table \ref{tab:1}, the robot's width is at its maximum (d = 202 \si{\milli\meter}) when the posture is set to 0\si{\degree}. Conversely, at a posture of -60\si{\degree}, the robot's width is at its minimum (d = 150 \si{\milli\meter}). The robot attempted to enter the narrower channel (width = 195 \si{\milli\meter}) after walking in a wider channel (width = 330 \si{\milli\meter}) with a fixed posture of 0\si{\degree}. However, this experiment was unsuccessful, as the robot collided with the side walls upon entering the narrower channel (SI movie- 1:06 to 1:16). Following the collision, the robot either remained stationary, attempting to enter the narrow channel or lost its stability and fell down. Alternatively, the robot was tested by initially moving through a broader channel with a fixed posture of 0\si{\degree}, and upon reaching the entry point of the narrower channel after completing one full cycle, the robot adjusted its posture from 0\si{\degree} to -60\si{\degree}. Subsequently, the robot moved inside the narrower channel at a posture of -60\si{\degree} (SI movie- 0:48 to 1:05). This sequence of actions was replicated five times, and in each repetition, the experiment resulted in a consistently successful outcome. It is to be noted that the robot does not make use of any obstacle-detecting sensor. Hence, while carrying out the experiments, the robot is placed at a pre-defined distance before the obstacle and the robot program is written in a way that the robot changes its posture after completing the pre-defined distance.

\subsection{Low-Clearance Avoidance}

Similar to the last experiment where the robot moved through a narrow channel, in this experiment, the robot goes through a tunnel of low height (200 \si{\milli\meter}) by avoiding collision with the roof of the tunnel. As can be seen in Table \ref{tab:1}, the maximum height of the robot is 240 \si{\milli\meter} and it attains this height when it has a posture of -60\si{\degree}. At 40\si{\degree} posture, the robot's height is least at 180 \si{\milli\meter}. The robot walks at -60\si{\degree} posture before entering the tunnel. In the experiments, the robot with -60\si{\degree} posture completes two cycles and stops at the entry point of the tunnel, as shown in Fig. \ref{fig10}.  At the entry point of the tunnel, the robot changes its posture from -60\si{\degree} to 40\si{\degree}. In doing so, the robot's height gets reduced from 240 \si{\milli\meter} to 180 \si{\milli\meter}. Then, the robot at 40\si{\degree} posture moves inside the tunnel fully by completing two cycles (SI movie- 1:17 to 1:35). This experiment is also carried out five times. Without the posture change, the robot collides with the tunnel, loses its stability, and falls (SI movie- 1:36 to 1:45). This experiment also highlights the importance of posture change in averting head-on collisions. 

\begin{figure}[!t]
\centering
\includegraphics[width=\columnwidth]{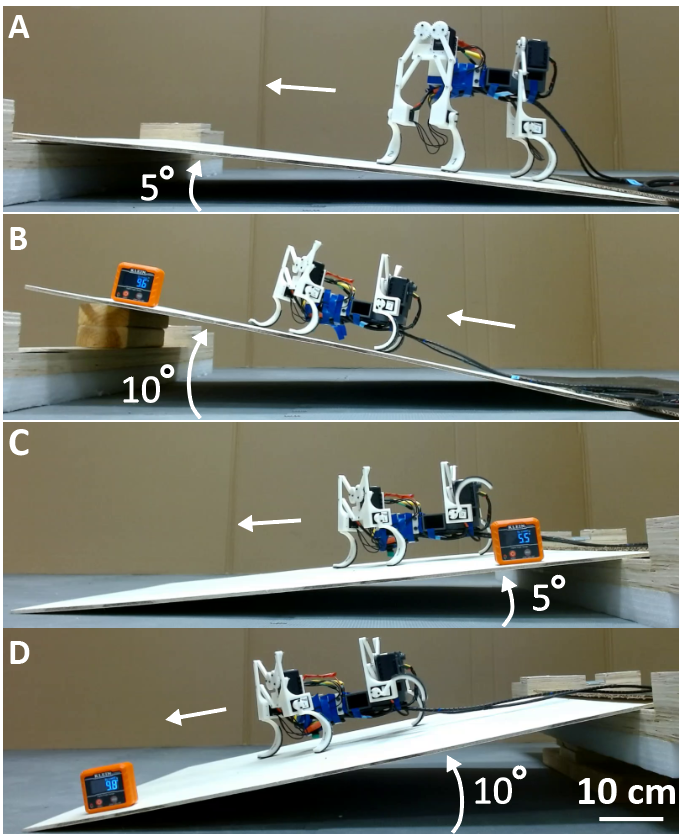}
\caption{\textbf{Successful walking on a ramp with different postures}: (A) (5\si{\degree} inclination uphill) with -60\si{\degree} posture and 70\si{\degree} body angle. (B) (10\si{\degree} inclination uphill) with 40\si{\degree} posture and 60\si{\degree} body angle. (C) (5\si{\degree} inclination downhill) with 40\si{\degree} posture and 40\si{\degree} body angle. (D) (10\si{\degree} inclination downhill) with 40\si{\degree} posture and 40\si{\degree} body angle. The arrows show the direction of motion of the robot.}
\label{fig12}
\end{figure}
\subsection{Obstacle Avoidance}

This experiment attempts to show how changing the robot's posture can be an effective way to overcome obstacles of different sizes. By adjusting its posture, the robot can adapt to the shape and size of the obstacle on its path. It might not always be possible to circumvent obstacles, especially when the robot passes through a narrow tunnel with not enough space for the robot to go around. Hence, these experiments further validate the importance of adaptable morphology in robots. With posture change, the robot goes past the obstacle without changing its path.

In these experiments, the robot traverses a level ground surface encountering three randomly sized obstacles that are smaller than the maximum clearance of the robot body, as shown in Fig. \ref{fig11}. The first obstacle is a cube with an edge length of 38 \si{\milli\meter}. The second obstacle is also a cube with an edge length of 50 \si{\milli\meter} and is placed at a distance of 300 \si{\milli\meter} from the first obstacle. The third obstacle is a cuboid with a length of 100 \si{\milli\meter}, width and height each of 50 \si{\milli\meter}. It is placed at a distance of 350 \si{\milli\meter} from the second obstacle. The robot is placed at a distance of 200 \si{\milli\meter} ahead of the first obstacle. Due to the absence of obstacle-detecting sensors in the current version of the robot, the relative distance between the obstacles is manually selected to allow sufficient time for the robot to traverse each obstacle consecutively.

\begin{figure}[!t]
\centering
\includegraphics[scale=0.7]{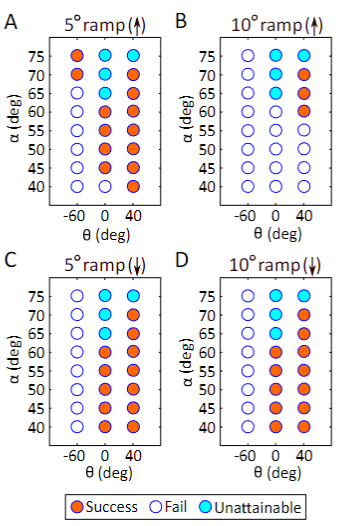}
\caption{\textbf{Performance of the robot on inclines and declines.} climbing (A)  upwards on a 5\si{\degree} ramp, (B)  upwards on a 10\si{\degree} ramp, (C)  downwards on a 5\si{\degree} ramp, (D) downwards on a 10\si{\degree} ramp}
\label{fig13}
\end{figure}
At the beginning of the experiment, the robot is at 40\si{\degree} posture. From Table \ref{t4}, we know that the robot body's height from the ground, h = 25 \si{\milli\meter} when the robot is at 40\si{\degree} posture. So, the robot after completing one cycle, stops in front of the second obstacle and changes its posture to 20\si{\degree}. At 20\si{\degree} posture, h$=$40 \si{\milli\meter}. The robot at 20\si{\degree} posture overcomes the first obstacle in two cycles and stops in front of the second obstacle. It changes its posture to -40\si{\degree} at which h$=$95 \si{\milli\meter}. The robot at -40\si{\degree} posture overcomes the second obstacle in two cycles and stops in front of the third obstacle. It changes its posture to 0\si{\degree} at which h$=$65 \si{\milli\meter}. To overcome the third obstacle, the robot could have moved with -40\si{\degree} posture but at -40\si{\degree} posture, the distance between the legs is 140 \si{\milli\meter}. In order to increase the distance between the legs, the robot's posture is changed to 0\si{\degree} in which the distance between the legs is 162 \si{\milli\meter}. The robot at 0\si{\degree} posture overcomes the third obstacle by completing two cycles (SI movie-  1:46 to 2:17). Note that, the motion of the robot is open-loop, i.e. it was programmed to change its postures within a specific time step. This experiment is carried out five times with a successful outcome each time.

In order to demonstrate the necessity of posture changes for the robot to successfully navigate the obstacles, a supplementary set of control experiments was conducted. At 40\si{\degree} posture, the robot body strikes the first obstacle. After that, upon encountering the obstacles, the robot's body remains stationary in its current position, either attempting to move forward or becoming immobilized with its legs suspended in the air, without establishing contact with the ground surface (SI movie- 2:18 to 2:32). In another experiment, the robot overcomes the first obstacle at 20\si{\degree} posture and then tries to overcome the second obstacle without changing its posture. Similar to the prior case, following the robot body's collision with the second obstacle, it exhibits behavior wherein the body remains stationary in an attempt to proceed, or it becomes immobilized atop the obstacle, with its legs suspended in the air and no ground contact established. 

Hence, the posture change mechanism empowers the quadruped robot to dynamically adjust its body height above the ground and the inter-leg spacing in direct response to encountered obstacles within its trajectory. This ability to modulate posture facilitates enhanced adaptability and maneuverability when navigating through complex environments and contributes to the robot's agility and efficiency in negotiating obstacles effectively, thereby expanding its range of potential applications across diverse terrains and scenarios.

\begin{figure}[!t]
\centering
\includegraphics[width=\columnwidth]{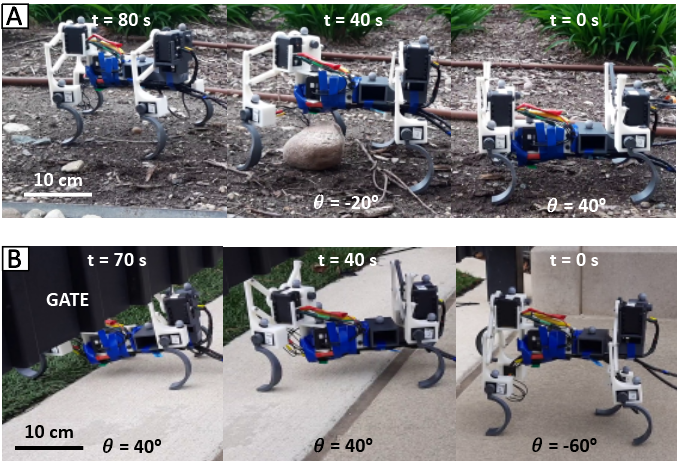}
\caption{\textbf{Robot motion in outdoor environment} (A) Overcoming a big rock on its path by changing posture from +40\si{\degree} to -20\si{\degree}, (B) Going inside the gate of a building by changing its posture from -60\si{\degree} to +40\si{\degree}}
\label{fig15}
\end{figure}
\subsection{Ramp Traversal}

In this experiment, the robot's stability is analyzed when it walks, both uphill and downhill, on 5\si{\degree} and 10\si{\degree} ramps. This experiment was considered important as in the previous subsections, the robot mostly moved on a flat surface while circumventing tunnels and obstacles through posture change. However, in real-world surroundings, the robot may need to walk through inclines and steep terrains with tunnels and rocks. 

\subsubsection{Uphill}
The robot walks for four cycles on a 5\si{\degree} ramp with posture angle, $\theta=$40\si{\degree} and amplitude of body angle, $\alpha=$40\si{\degree} without falling. However, with 0\si{\degree} posture and 40\si{\degree} amplitude, the robot is able to complete two full cycles and falls at the start of the third cycle (SI movie- 3:30 to 3:39). For -60\si{\degree} posture and 40\si{\degree} body angle, the robot is unable to complete even one cycle and falls before that. Each of these experiments was carried out at least five times and similar observations were made each time. 

In order to make the robot move for full four cycles with 0\si{\degree} and -60\si{\degree} postures on a 5\si{\degree} ramp, $\alpha$ of the robot is increased to 45\si{\degree}. With 45\si{\degree} amplitude, the robot completes full four cycles on a 5\si{\degree} ramp both with 0\si{\degree} (SI movie- 3:39 to 3:54) and 40\si{\degree} postures. However, with $\alpha=$45\si{\degree} and $\theta=$-60\si{\degree}, the robot is unable to complete even one cycle and falls. In order to make the robot move on a 5\si{\degree} ramp with $\theta=$-60\si{\degree}, $\alpha$ is sequentially increased from 45\si{\degree} to 75\si{\degree} with an increment of 5\si{\degree}. The robot at $\theta=$-60\si{\degree} kept falling in the first cycle itself for different values of $\alpha$ except when $\alpha=$70\si{\degree}. With $\theta=$-60\si{\degree} and $\alpha=$70\si{\degree}, the robot could walk for a full four cycles on the 5\si{\degree} ramp without falling (SI movie- 2:33 to 2:47). 

Next, the robot walks on a 10\si{\degree} ramp for four cycles. With $\theta=$40\si{\degree} and $\alpha=$40\si{\degree}, the robot is unable to climb and falls without completing a single cycle. Hence, $\alpha$ is increased with an increment of 5\si{\degree}. With $\theta=$40\si{\degree} and $\alpha=$60\si{\degree}, the robot is able to climb on a 10\si{\degree} ramp for four cycles (SI movie- 2:48 to 3:01). However, with $\theta=$0\si{\degree} and $\alpha=$60\si{\degree}, the robot is able to complete only one cycle and falls in the second cycle. With -60\si{\degree} posture, the robot is not able to complete even one cycle and falls irrespective of the $\alpha$ value. 

In this section, it is important to highlight that when the robot undergoes a posture change, it becomes notably less stable, a phenomenon that becomes especially pronounced when navigating inclines as opposed to flat terrain. Therefore, the experiments conducted above demonstrate that, during incline traversal, enhancing the amplitude of the body angle across all body postures can significantly improve the robot's stability. The performance of the robot in terms of its ability to complete full four cycles during uphill movement for different postures and body angles is shown in Fig.\ref{fig13} A and B. In Fig.\ref{fig13}, ''success`` denotes the robot's ability to complete all four cycles, while ''Fail``  indicates a scenario in which the robot's legs do not interfere with other body parts, however, it is unable to successfully complete all four cycles. ''Unattainable`` refers to the configuration of the robot where the robot legs start interfering with other legs or the robot body and the motion gets locked. It may be noted that with $\theta=$40\si{\degree}, the robot can have a maximum $\alpha$ of 70\si{\degree}. Similarly, the maximum $\alpha$ for $\theta=$0\si{\degree} and $\theta=$-60\si{\degree} is 60\si{\degree} and 75\si{\degree}, respectively. If $\alpha$ gets bigger than its maximum value, the robot reaches an unattainable or singular configuration. These experiments also show that during uphill movement, the stability of the robot also depends on the height of the robot body from the ground, h \ref{fig3}. The lower the value of h, the greater the stability of the robot. 

\subsubsection{Downhill}

The robot walks on a 5\si{\degree} ramp for three cycles. For 40\si{\degree} and 0\si{\degree} postures, the robot completes 3 cycles without falling for all the body angles, as shown in Fig. \ref{fig13}. The locomotion of the robot with 40\si{\degree} posture and 40\si{\degree} body angle can be found in SI movie- 3:01 to 3:16). As the robot moves down the ramp, it has a tendency to fall towards the front irrespective of the posture and the body angle. This phenomenon gets more pronounced as the robot body's height increases from the ramp surface. Thus, for -60\si{\degree} posture, the robot falls on its front by tipping over rather than falling on the back side which is observed when the robot climbs up the ramp (SI movie- 3:55 to 4:01). With -60\si{\degree} posture, the robot fails to complete one cycle and falls irrespective of the body angle. However, with -40\si{\degree} posture, the robot climbs down the ramp for all the body angles. The locomotion of the robot with 0\si{\degree} posture and 45\si{\degree} body angle can be found in SI movie- 4:02 to 4:11.

Next, the robot walks on a 10\si{\degree} ramp for three cycles. Like in the previous case, the robot completes 3 cycles without falling for all the body angles for 40\si{\degree} and 0\si{\degree} postures. For -60\si{\degree} posture, the robot falls on its front by tipping over and is unable to complete one cycle. However, the robot is able to complete three cycles with -20\si{\degree} postures for all the body angles. The locomotion of the robot with 40\si{\degree} posture and 40\si{\degree} body angle can be found in SI movie- 3:17 to 3:27.  

\begin{table}[!t]
\centering
\caption{Suitable postures of the robot while moving in an unstructured environment}
\label{t5}

\begin{tabular}{ c c c }
\hline
\hline
Unstructured  &  \multirow{2}{*}{Posture Range}  &  \\

Environment &   \multirow{2}{*}{(in degrees)}     &   Summary         \\

Type &                &            \\

\hline 

Narrow   & \multirow{3}{*}{-60 to -40}    &    The robot's width (d) is minimum  \\

Tunnel   &  &  when it attains a posture between\\

Traversal   &  &   -40 and -60 degrees  \\

   &   &    \\

Low  & \multirow{3}{*}{20 to 40}    &    The robot's height (H) is minimum \\

Clearance   &  &  when it attains a posture between\\

Avoidance   &  &   +20 and +40 degrees  \\

   &    &       \\

\multirow{3}{*}{Obstacle} & \multirow{4}{*}{-60 to 0}  &  The robot's ground clearance (h)  \\
\multirow{3}{*}{Avoidance}   &   & and width (d) are greatest when   \\
 &   &  it attains a posture of \\

   &   & -60 and 0 degrees, respectively \\

      &    &       \\
      
\multirow{3}{*}{Ramp}  & \multirow{3}{*}{0 to 40} &  The robot is mostly stable when \\

     &    &    it climbs with a posture between   \\

        &    &    0 and 40 degrees   \\
        
\hline
\end{tabular}
\label{tab:2}
\end{table}

\subsection{Outdoor experiments}

The adaptable nature of the robot body also helps it to overcome and circumvent obstacles in outdoor settings without the use of any obstacle-detecting sensor. In Fig. \ref{fig15} A, the robot is able to overcome a big rock on its path by changing its posture (SI movie- 4:43 to 5:00). Conversely, when the posture change is not applied, the robot becomes immobilized and remains stuck on the obstacle (SI movie- 5:01 to 5:12). Here, the robot changes its posture from +40\si{\degree} to -20\si{\degree}. In Figure \ref{fig15} B, the robot demonstrates its ability to navigate through a building entrance by adopting a posture change from -60\si{\degree} to +40\si{\degree}  (SI movie- 4:13 to 4:36). Without the posture change, the robot experiences a collision with the gate, resulting in a subsequent fall (SI movie- 4:37 to 4:42). These two experiments validate the importance of posture change in quadruped robots to adapt themselves to real-world environments. It may be noted that since the surfaces in outdoor settings are found to be very rough, the outdoor studies are conducted on legs without any rubber patches.

\section{Conclusion}
\label{secconc}

The paper describes a quadruped robot called Adjustbot, which takes inspiration from crocodiles to achieve variable posture and adaptability to unstructured environments. The robot's posture change is accomplished through two parallelogram mechanisms, actuated simultaneously by a single servo through a gear drive. A bio-inspired approach is utilized to analyze the stability and gait selection of the robot. 

After the design, prototyping, stability analysis, and gait selection, the robot's posture-changing ability is tested in indoor and outdoor environments. The primary goals of the experiments are to avoid collisions, navigate through narrow channels, and overcome obstacles, and inclines. The robot achieves this by employing dynamic posture changes and lateral undulation, which enables it to adapt to challenging terrains. One notable aspect of the work is that it aims to demonstrate the robot's robustness and adaptability without relying on learning algorithms or posture-maintaining sensors like Inertial Measurement Units (IMUs). Overall, the experiments emphasize the importance of the robot's ability to change its morphology (shape and posture) to overcome hurdles in dynamically challenging environments, making it more adaptable and stable without the need for complex learning or sensing mechanisms. Through ramp experiments, it is also found that the body angle's amplitude directly affects the robot's stability. The stability of the robot increases proportionally with higher amplitudes.

In our previous study, we investigated the impact of a variable stiffness robotic tail on the performance of a sprawling quadruped robot, emphasizing improved stability, maneuverability, and climbing ability \cite{buckley2023effect}. The addition of a bio-inspired, flexible tail to Adjustbot will be the focus of future studies in order to improve the robot's stability when it navigates slopes and other irregular terrain. On top of that, additional research will be done on the created robot for other field applications, such as burrowing and swarm applications. \section{Acknowledgment}
We would like to express our appreciation to the members of Notre Dame MiNiRo-Lab for their invaluable contributions and insightful discussions.
\bibliographystyle{ieeetr}

\end{document}